\documentclass{article} % For LaTeX2e
\usepackage{iclr2025_conference,times}

\usepackage{amsmath,amsfonts,bm}

\def\eqref#1{equation~\ref{#1}}
\def\1{\bm{1}}

\DeclareMathAlphabet{\mathsfit}{\encodingdefault}{\sfdefault}{m}{sl}
\SetMathAlphabet{\mathsfit}{bold}{\encodingdefault}{\sfdefault}{bx}{n}

\usepackage{hyperref}
\usepackage{url}

\usepackage{booktabs}
\usepackage{graphicx}
\usepackage{color, colortbl}
\usepackage{enumitem}

\usepackage{verbatim}
\usepackage{mdframed}

\usepackage{inconsolata}
\newcommand{\model}{\textsc{SiReRAG}}
\title{\model{}: Indexing Similar and Related Information for Multihop Reasoning}

\author{\stepcounter{footnote}Nan Zhang$^{\clubsuit}$\thanks{Work done while interning at Salesforce AI Research.} \quad Prafulla Kumar Choubey$^\diamondsuit$ \quad Alexander Fabbri$^\diamondsuit$ \quad Gabriel Bernadett-Shapiro$^\diamondsuit$\\ {\bf Rui Zhang$^{\clubsuit}$ \quad Prasenjit Mitra$^{\clubsuit}$ \quad Caiming Xiong$^\diamondsuit$ \quad Chien-Sheng Wu$^\diamondsuit$}\\
$^{\clubsuit}$The Pennsylvania State University \quad $^\diamondsuit$Salesforce AI Research \\
\texttt{\{njz5124,rmz5227,pmitra\}@psu.edu}\\ \texttt{\{pchoubey,afabbri,gbernadettshapiro,cxiong,wu.jason\}@salesforce.com}
}

\iclrfinalcopy % Uncomment for camera-ready version, but NOT for submission.
\begin{document}

\maketitle

\begin{abstract}

Indexing is an important step towards strong performance in retrieval-augmented generation (RAG) systems. However, existing methods organize data based on either semantic similarity (similarity) or related information (relatedness), but do not cover both perspectives comprehensively. 
% Related information includes signals such as shared entities and topics.
Our analysis reveals that modeling only one perspective results in insufficient knowledge synthesis, leading to suboptimal performance on complex tasks requiring multihop reasoning.
% As our analysis reveals, modeling only one perspective leads to insufficient knowledge synthesis and suboptimal performance on complex tasks requiring multihop reasoning.
% , due to the limited coverage of supporting information with each perspective and their divergent nature. % nature of these two perspectives.
In this paper, we propose \model{}, a novel RAG indexing approach that explicitly considers both similar and related information. 
On the similarity side, we follow existing work and explore some variances to construct a similarity tree based on recursive summarization.
% We first investigate the hierarchical structure of an indexing tree and find a more structured tree design would not offer significant performance improvement. We thus group similar information via recursive summarization to construct our similarity tree. 
On the relatedness side, \model{} extracts propositions and entities from texts, groups propositions via shared entities, and generates recursive summaries to construct a relatedness tree. We index and flatten both similarity and relatedness trees into a unified retrieval pool. Our experiments demonstrate that \model{} consistently outperforms state-of-the-art indexing methods on three multihop datasets (MuSiQue,  2WikiMultiHopQA, and HotpotQA), with an average 1.9\% improvement in F1 scores. As a reasonably efficient solution, \model{} enhances existing reranking methods significantly, with up to 7.8\% improvement in average F1 scores. Our code is available at \url{https://github.com/SalesforceAIResearch/SiReRAG}.

\end{abstract}

% Inserted by Nan
% \fancyhead{}
\section{Introduction}

Retrieval-augmented generation (RAG) has shown strong potential in augmenting large language models (LLMs) with highly specialized and constantly updated knowledge~\citep{NEURIPS2020_6b493230,gao2024retrievalaugmentedgenerationlargelanguage}. Getting rid of fine-tuning LLMs, it is an efficient method for handling users' queries that require domain knowledge. A typical RAG pipeline may involve chunking, embedding, indexing, retrieval with queries, reranking, and LLM response generation~\citep{wang2024searchingbestpracticesretrievalaugmented}.

% \jw{cite this work: https://arxiv.org/pdf/2407.01219v1}.

The indexing step is a prerequisite. It focuses on organizing a large amount of data and serves as an upstream step of retrieval. For example, RAPTOR~\citep{sarthi2024raptor} shows a significant performance improvement by adding recursive summaries to text chunks of a dataset, which demonstrates the potential of adding synthesized information for retrieval. The added recursive summaries combine semantically similar information within a dataset. GraphRAG~\citep{edge2024local}, on the other hand, indices an entire corpus via an entity-guided knowledge graph. It then constructs summaries from closely-related entities and their mentions, synthesizing the connections and relatedness among different pieces of information.

\begin{figure}[t!]
    \centering
    % \vspace{0.3cm}
    \includegraphics[width=\textwidth]{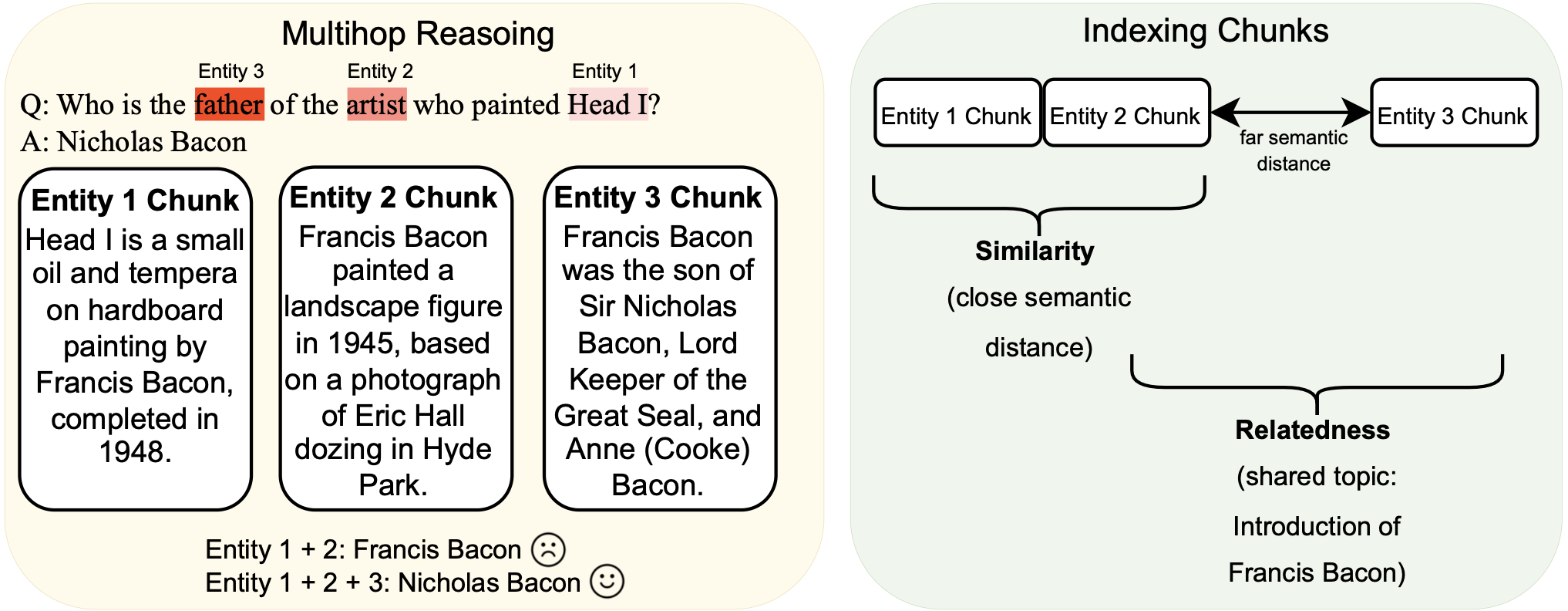}
    \caption{\textbf{Challenges of existing RAG indexing methods for multihop reasoning.} Entity 1 and 2 chunks contain similar information while entity 2 and 3 chunks contain related contents. Since synthesizing information only based on entity 1 and 2 (or entity 2 and 3) will lead to a higher probability of a wrong answer, an indexing method that considers both similarity and relatedness is needed to maximize retrieving relevant knowledge for multihop questions.}

    \label{fig:motivation}
    \vspace{-0.3cm}
\end{figure}

However, none of the existing methods address the importance of indexing from both similarity and relatedness sides, which limits a holistic understanding of the provided dataset. We define \textbf{similarity} as the semantic distance of text pieces and \textbf{relatedness} as the degree of connection of texts based on signals such as entities and %factoids
propositions. 
Indexing similar and related information facilitates more comprehensive knowledge integration than indexing individual kind of information. As shown in Figure~\ref{fig:motivation}, a complex question that involves two hops of reasoning requires the retrieval and synthesis of relevant entity chunks. For example, synthesizing entity 1 and 2 chunks would encourage LLMs to generate ``Francis Bacon'', which is a common mistake. Entity 1 and 2 chunks have close semantic distance. 
On the other hand, synthesizing entity 2 and 3 chunks would not maximize the probability of the correct answer even when entity 1 chunk is retrieved, since LLMs may struggle to reason about the painter of ``Head I'' in long-context environment~\citep{liu2024lost}. In this case, entity 2 and 3 chunks are related due to a shared topic.
% \jw{sentences after here in this paragraph did not add any new or meaningful info to this paper, I would suggest you use a concrete example in Fig 1 to show this motivation. I suggest you want some fake chunks in the figure like hop 1 chunk: "... artist Francis Bacon painted his father's head ...", hop 2 chunk: "... Nicholas talked to head of art department...", and hop 3 chunk: "... XXX created Head I in 19XX; XXX talked to his father Nicholas Bacon; ..." you can decide what to write but with an example it can help you better describe and motivate SireRAG.}
Therefore, it is important to synthesize
% \jw{why use the word "synthesize"? we are not building QA datasets?} 
both similar and related information in order to maximize the chance of retrieving relevant knowledge, in particular, for multihop reasoning questions.
% , since this type of complex questions requires comprehensive knowledge synthesis across multiple hops of %facts information pieces. 
We verify the bottleneck of solely modeling similarity or relatedness through quantitative methods in Section~\ref{sec:bottleneck} and demonstrate that neither perspective yields the optimal performance.
% Without explicitly considering both similarity and relatedness, it would be challenging to maximize the retrieval performance of RAG pipelines. \jw{you can also mention the conclusion in Section 3 here.}

\begin{figure}[t!]
    \centering
    % \vspace{0.3cm}
    \includegraphics[width=\textwidth]{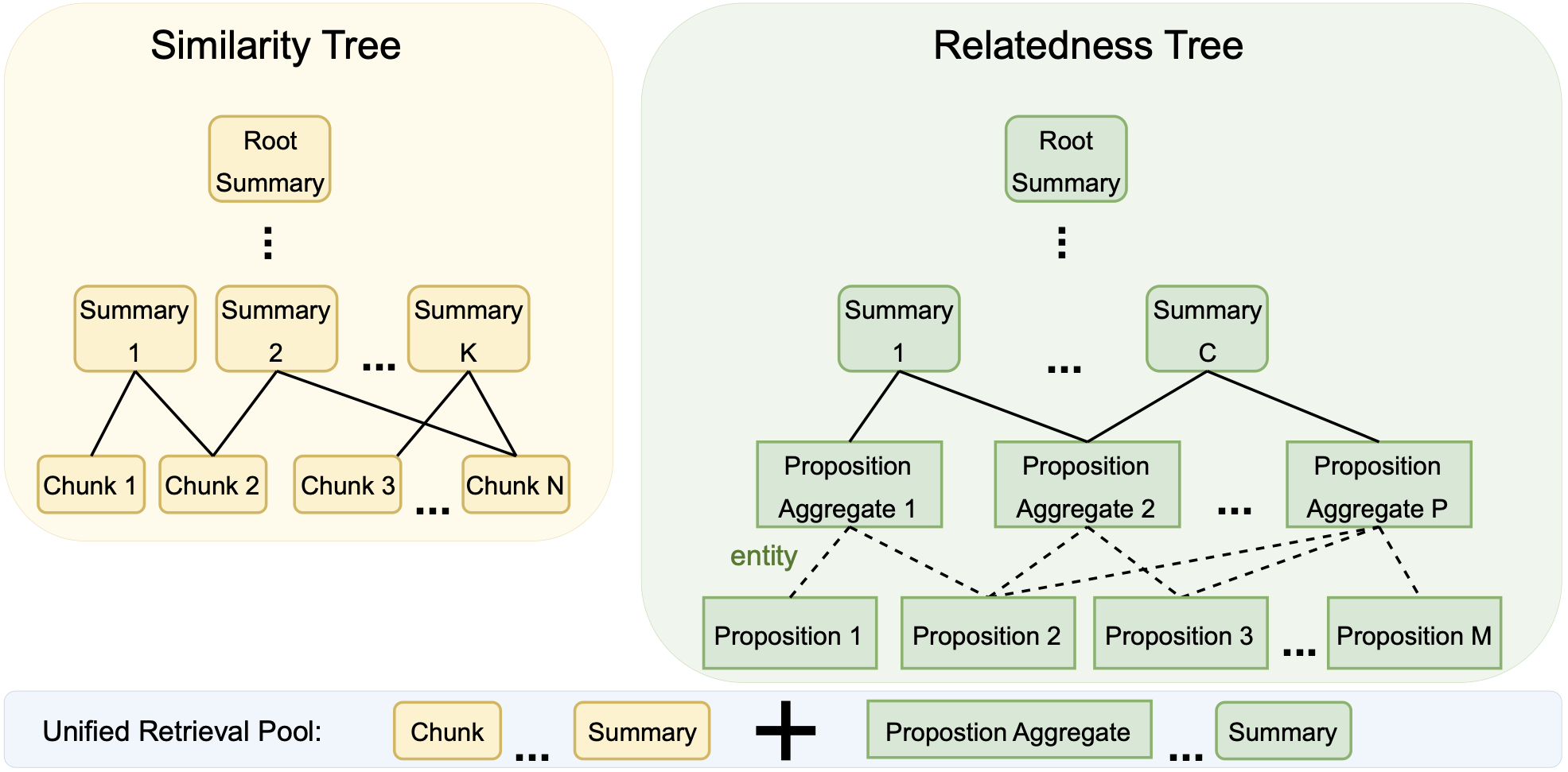}
    \caption{\textbf{\model{} Tree.} We adopt RAPTOR~\citep{sarthi2024raptor} to construct the similarity tree (left). On the right, we construct the relatedness tree by clustering the propositions based on their entities to get proposition aggregates and having recursive summaries on top. Note that propositions are not included in the relatedness tree, so their connections to proposition aggregates are marked with dashed lines. 
    % \jw{facts to be updated with propositions, and maybe making related trees taking 70\% of space and similarity 30\% only? given that left-hand side is not sth we proposed.}
    % \jw{in your best method, did you flatten both tress and combine all the chunks/summaries? If so, let's add one layer on the top to indicate that.}
    }
    \label{fig:pipeline}
    \vspace{-0.3cm}
\end{figure}

In this paper, we propose \model{}, which stands for RAG indexing of \underline{\textbf{si}}milarity and \underline{\textbf{re}}latedness as shown in Figure \ref{fig:pipeline}. On the similarity side, \model{} follows RAPTOR~\citep{sarthi2024raptor} to build a recursive tree based on chunk similarity. We adopt a shallow tree with 4 levels in total.
% we use ... (add some details of hyper-parameter selection).
On the relatedness side, \model{} first extracts entities (\emph{e.g.}, ``Sonnet 110'' and ``William Shakespeare'') and fine-grained propositions (\emph{e.g.}, ``Sonnet 110 is one of 154 sonnets written by William Shakespeare.'') from each text chunk/document using LLMs. We group these propositions into aggregated ones via entities, simply concatenating them with the original order in chunk/document. These proposition aggregates contain related information, because they mention shared entities. Then, recursive summaries with soft clustering are built on top of those aggregated propositions. Finally, we index both trees by flattening nodes in each tree for retrieval. 
% \model{} performs clustering based on both similarity and relatedness within corpus-level data. 
% For efficiency and effectiveness, we first decide to adopt Raptor's tree structure~\citep{sarthi2024raptor} and explore whether a more structured design of trees would be beneficial. After finding that organizing texts into different levels of abstraction (\emph{e.g.}, a hierarchical structure) does not provide a significant improvement, we perform clustering on a flat structure of text chunks to find semantically close information. \jw{this part is ablation study, which is not important to mention here.}
% Then, to emphasize relatedness, we extract entities and fine-grained factoids from a dataset and aggregate these factoids via entities. Recursive summaries are built on top of the clustered chunks and the fact aggregates to mimic the tree structure of Raptor, which yields two separate trees. 

We show that \model{} is effective on a variety of multihop question answering (QA) datasets including MuSiQue~\citep{trivedi2022musiquemultihopquestionssinglehop}, 2WikiMultiHopQA~\citep{ho-etal-2020-constructing}, and HotpotQA~\citep{yang-etal-2018-hotpotqa}. 
We find \model{} consistently outperforms the strongest RAG indexing methods, achieving an average F1 improvement of at least 1.9\%.
% \jw{that's say up to X\% instead of at least}. 
% \jw{TODO} 
We conduct an ablation study on several components in \model{} and observe that adding proposition aggregates to the similar information within a dataset yields the most improvement, which echoes our motivation. \model{} is also a reasonably efficient model, as it does not introduce many lengthy or redundant retrieval candidates.
% XXX (write the main takeaway of the finding, for example, XXX part is the most crucial to overall performance).
% \jw{reviewers might also ask what is the inference speed/memory cost of your method compare to raptor, let's add this analysis and briefly mention here.}
% Through combining various text granularities, our contribution is the idea of indexing with the consideration of both similarity and relatedness to maximize the probability of retrieving correct answers, which fills the gap existing methods. 
% An ablation study is conducted to demonstrate the utility of each component of \model{} and showcase the efficiency of our design. 

\section{Related Work}
\paragraph{RAG} RAG is a framework that integrates retrieval mechanisms into generative models to enhance text generation by leveraging external knowledge. This concept has evolved from earlier retrieval-based methods such as DrQA~\citep{chen-etal-2017-reading} and DPR~\citep{karpukhin-etal-2020-dense}. Instead of separating retrieval and generation phases, researchers also showed the potential of tightly coupling retrieval and generation into an end-to-end framework~\citep{NEURIPS2020_6b493230}. 

Recent advances in retrieval mechanisms include leveraging LLMs as retrievers~\citep{yu2023generateretrievelargelanguage,sun2023recitationaugmentedlanguagemodels} and exploring retrieval granularity such as proposition~\citep{chen2023dense}. Here each proposition is an atomic expression that contains a factoid presented in natural language, which is similar to the propositions used by \model{}. Inspired by the idea of retrieval granularity, we combine several different text granularities (entity, propositions, text chunk, and summary) to index data. As a representative work on text segmentation, researchers proposed the frst supervised approach to generate hierarchical segmentation structures~\citep{nair-etal-2023-neural}.

\paragraph{RAG Indexing} An earlier work~\citep{NEURIPS2020_6b493230} shows the benefits of document index on the overall retrieval performance. Most recent works on RAG indexing include RAPTOR that builds a tree with recursive summaries~\citep{sarthi2024raptor}, HippoRAG that leverages the hippocampal indexing theory of human long-term memory for deep knowledge integration~\citep{gutierrez2024hipporag}, and GraphRAG that constructs an entity-guided knowledge graph~\citep{edge2024local}. However, all these works overlooked the importance of considering both similarity and relatedness during indexing. Specifically, RAPTOR integrated knowledge only based on similarity, and the other two only considered relatedness to synthesize information. Although these three approaches achieved competitive performance on different kinds of datasets and HippoRAG has the same goal as ours (multihop reasoning benchmarks), \model{} is fundamentally different from them in terms of the explicit incorporation of both similar and related knowledge.

% \jw{The related work looks pretty thin to me. Can we check how Raptor wrote theirs can expand here?}

\section{Bottleneck of Solely Modeling Similarity or Relatedness}
\label{sec:bottleneck}
To verify our hypothesis of insufficient knowledge integration when solely modeling similarity or relatedness, we perform different kinds of clustering philosophies on a retrieval corpus of MuSiQue~\citep{trivedi2022musiquemultihopquestionssinglehop}. Using the same retrieval corpus as HippoRAG~\citep{gutierrez2024hipporag}, we obtain 1000 questions from the validation set of MuSiQue along with their candidate passage clusters (each cluster includes supporting and distractor passages). We include distractor passages for a more realistic setting, since they are semantically close and/or related to the supporting candidates. Treating these clusters as the gold labels for different queries, we run our own clustering on all passages based on either similarity or relatedness.
% \jw{we need to explain why we treat distractor passages also as gold labels, it is because those distractors also have the same entities mentioned?}

Following RAPTOR~\citep{sarthi2024raptor}, we use Gaussian Mixture Models (GMMs) to perform soft clustering, assuming that a candidate passage can belong to multiple clusters. 
For similarity, we run GMMs on the deep representations of all passages to find semantically similar groups. For relatedness, we first extract the topic of each passage using OpenAI GPT-4o
% \jw{in free-form text} 
and then cluster passages based on the representations of their topics, because we assume that related passages share similar topics. To obtain deep representations of either passages or topics, we use OpenAI text-embedding-3-small for a balance of performance and efficiency.
% We perform GMMs on the vector embeddings of these topics. 
% \jw{We use X embeddings in our experiments because ...}
% To evaluate the overlapping ratio with gold labels
To evaluate the overlapping ratio among clusters, we convert every cluster into pairwise connections. For instance, given two clusters \texttt{[1, 2, 3]} and \texttt{[3, 5]}, the resulting pairwise connections are as follows: ``\texttt{1-2}'', ``\texttt{1-3}'', ``\texttt{2-3}'', and ``\texttt{3-5}''. By computing the number of shared pairwise connections between gold clusters and predicted clusters, we aim to see how many pairwise connections from gold labels are covered by our two clustering philosophies, and we believe this coverage is a key indicator of knowledge integration for RAG indexing. We also report the overlapping ratio between ``supporting only'' similarity and relatedness to make a point of combining both clustering philosophies.

Table~\ref{tab:testing_similarity_relatedness} shows the coverage of supporting only or all passages as gold labels. Higher coverage is obtained when we only use supporting passages as gold, which indicates the commonality of supporting ones with respect to distractors. Both philosophies are able to capture this commonality. However, all these percentage scores are low, with the highest one being around one-fifth of the total coverage (19.14\%). This indicates a significant insufficiency in knowledge synthesis when we model similarity or relatedness solely. As a result, the chance of retrieving relevant knowledge for multihop reasoning questions could be suboptimal. 

% \jw{below passage is hard to digest for me so I make some changes, please make sure it is correct:}
Taking a further step in the supporting-only setting, we find that only 50.15\% of the correct similarity connections overlap with the correct relatedness connections. Correct similarity connections are those gold connections covered by similarity, while correct relatedness connections are those covered by relatedness. Conversely, 68.85\% of the correct relatedness connections overlap with the correct similarity connections.
% we observe that the overlap between ``supporting only'' similarity and relatedness accounts for around half of the retrieved gold passages when using only supporting passages (50.15\% of the 19.14\% gold supporting passages). 
In other words, predictions based on similarity or relatedness are not identical, and we can potentially leverage both to improve retrieval performance and facilitate a more comprehensive knowledge integration process.
% if we cluster information solely based on relatedness, we expect to cover only half of the passages retrieved by similarity. 
% However, this still highlights the divergent nature of the clustering outputs from our two philosophies, as neither overlapping score is close to 100\%. 
Although more customized clustering algorithms of each clustering philosophy can be proposed, combining both similarity and relatedness offers an effective and straightforward solution.
% \jw{Good trail on motivating this and I think this is a good experimental setup. But I think the most important analysis here is missing: What are the overlapping between 20.69 v.s. 14.69. What percentage of 14.69 is actually inside the 20.69? If the ratio is not close to 100\%, then we can make a point of combining.}

\begin{table*}[t!]
\caption{Coverage percentage between different clusters. MuSiQue clusters include supporting and distractor passages. The ``all'' setting treats both as gold, while ``supporting only'' uses only supporting passages as gold. We first show the coverage of supporting or all passages under two clustering philosophies. We then report the overlapping ratio between ``supporting only'' similarity and relatedness to motivate our work of combining both philosophies.
% We obtain the set intersection of ``supporting only'' similarity and relatedness. We then report the overlapping ratio of this intersection with respect to similarity and rela
% Coverage percentage of supporting or all passages under two clustering philosophies.
}
\label{tab:testing_similarity_relatedness}
\begin{center}
\resizebox{\textwidth}{!}{%
\begin{tabular}{@{}lcccccc@{}}
\toprule
& \multicolumn{2}{c}{Similarity Coverage} & \multicolumn{2}{c}{Relatedness Coverage} & \multicolumn{2}{c}{Overlapping Ratio} \\
\cmidrule(lr){2-3} \cmidrule(lr){4-5} \cmidrule(lr){6-7}
 & Supporting Only & All & Supporting Only & All & Overlap@Similarity & Overlap@Relatedness\\ \midrule
Coverage & 19.14\% & 10.70\% & 13.94\% & 8.51\% & 50.15\% & 68.85\%\\ \bottomrule
\end{tabular}
}
\end{center}
\end{table*}

\section{Methodology}
We propose \model{}, a RAG indexing framework guided by similarity and relatedness. As shown in Figure~\ref{fig:pipeline}, its left tree integrates information based on similarity while its right tree integrates information based on relatedness. As a first step, we study an alternative tree design to determine whether we can develop a generalized tree structure for similarity and relatedness trees, and beyond (Section~\ref{sec:hierarchy}). After the construction of the similarity tree, we extract propositions and their entities from our multihop reasoning dataset and perform clustering based on entities to synthesize related information (Section~\ref{sec:relatedness_method}). Indexing separate similarity and relatedness trees (Section~\ref{sec:index_both_method}), \model{} explicitly models both kinds of information within our dataset.

\subsection{Exploring a hierarchical structure of trees}
\label{sec:hierarchy}
For efficiency, we stick to a tree structure to organize texts and explore whether a more structured tree design would offer performance improvement. RAPTOR placed all text chunks at the bottom level and recursive summaries at upper levels, but this design does not closely follow the commonsense definition of a tree~\citep{zhang2002ontology}, where multiple levels of data abstraction are provided. The reason is that different text chunks of a document may showcase different levels of abstraction, which serves as the assumption of many document discourse trees~\citep{maekawa-etal-2024-obtain,liu2021dmrst}. For example, text chunks of the introduction section of a research paper are more likely to be more abstractive than those from the methodology section. Therefore, we first explore placing different text chunks on different levels of a tree based on their levels of abstractiveness.

For this analysis, we choose QuALITY dataset~\citep{pang-etal-2022-quality} instead of multihop reasoning ones, because the documents of QuALITY are longer, resulting in more text chunks than using the other datasets. We prompt GPT-4o to identify a two-level hierarchy for all candidate chunks: low (text chunks describing fine-grained details about a topic) and high (text chunks giving an overview of a topic and summarizing fine-grained details) abstractiveness. Summary nodes will still start from the second level (the level above the bottom level), so the second level will contain both summary nodes (of the low-abstractive chunks) and high-abstractive chunks. As in RAPTOR, nodes of the third level and above are recursive summaries of their children. Using the same text chunks across two methods, we compare RAPTOR against our hierarchical design. Since we simply retrieve the top 10 relevant nodes via an embedding model (\texttt{text-embedding-3-small}) when a query arrives, the idea of hierarchical structure of chunks would affect summary nodes due to the differences of their children.

\begin{table}[]
\centering
\caption{QA performance of having hierarchical text chunks on the validation set of QuALITY. Due to randomization of clustering and generation temperature of LLMs, we run each indexing method for 5 times and compute their average and standard deviation of accuracies.}
\label{tab:testing_hierarchy}
\begin{tabular}{@{}lcc@{}}
\toprule
 & Average Accuracy & Standard Deviation \\ \midrule
RAPTOR & 78.88 & 0.005 \\
Hierarchical Text Chunks & 78.76 & 0.004 \\ \bottomrule
\end{tabular}
\end{table}

% \begin{table}[]
% \centering
% \begin{tabular}{@{}lcccccc@{}}
% \toprule
% & \multicolumn{2}{c}{Similarity} & \multicolumn{2}{c}{Relatedness} \\
% \cmidrule(lr){2-3} \cmidrule(lr){4-5}
%  & Supporting Only & All & Supporting Only & All & Similarity $\rightarrow$ Relatedness & Similarity $\rightarrow$ Relatedness\\ \midrule
% Coverage & 19.14\% & 10.70\% & 13.94\% & 8.51\% & & \\ \bottomrule
% \end{tabular}
% \caption{Coverage percentage of supporting or all passages under two clustering philosophies.}
% \label{tab:testing_similarity_relatedness}
% \end{table}

% \begin{table}[]
% \centering
% \begin{tabular}{@{}lcc@{}}
% \toprule
%  & Average Accuracy & Standard Deviation \\ \midrule
% Raptor & 78.88 & 0.005 \\
% Hierarchical Text Chunks & 78.76 & 0.004 \\ \bottomrule
% \end{tabular}
% \caption{QA performance of having hierarchical text chunks on the validation set of QuALITY. Due to randomization of clustering and generation temperature of LLMs, we run each indexing method for 5 times and compute their average and standard deviation of accuracies.}
% \label{tab:testing_hierarchy}
% \end{table}

As shown in Table~\ref{tab:testing_hierarchy}, we do not observe a clear trend of improvement from using hierarchical text chunks, even averaged over 5 different runs. The RAPTOR tree is relatively more efficient, as identifying hierarchical text chunks and document discourse trees would require additional computation. Thus, we conclude that a more structured tree design would not significantly improve performance, as long as the correct information is indexed to answer a question. Since RAPTOR uses Gaussian Mixture Models and representations of text chunks to perform clustering, it sets a nice example of integrating knowledge based on similarity. We adopt it to construct our similarity tree as shown in the left part of Figure~\ref{fig:pipeline}. 

\subsection{Synthesizing information based on relatedness}
\label{sec:relatedness_method}
For relatedness, we need to synthesize information based on a different philosophy than RAPTOR. Because related information pieces always share some degree of connection (\emph{e.g.}, overlapping subjects), we assume that two text pieces are related if they mention the same entity (e.g., person, location, product, etc). %\jw{
For example, entity 2 and 3 chunks in Figure~\ref{fig:motivation} are unlikely to be clustered in the same group based on similarity, but since they both mention Francis Bacon, we are able to connect them together.
%}

\paragraph{Modeling Relatedness with Entity-Specific Propositions:}
% \pk{
To effectively use entities for organizing related content, we first need to determine the appropriate granularity for text pieces. There are three main limitations with directly connecting entities to standard text chunks. First, a chunk often contains information beyond the scope of a specific entity, making it challenging to localize information about one entity, potentially adding noise. Second, aggregating all chunks in an indexing corpus for each entity can result in hundreds of thousands of tokens for each entity, which may lead to long context performance issues, such as losing critical information in the middle~\cite{liu2024lost} or experiencing low coverage and citation performance~\cite{laban2024summary}. Third, linking with chunks will introduce redundancy as each chunk may be a part of multiple entity clusters. Therefore, inspired by recent works on retrieval granularity~\citep{liu2023towards,chen2023dense}, we propose to use short entity-specific ``propositions'' to represent fine-grained knowledge about entities and build our relatedness tree. %\jw{make sure propositions should be mentioned earlier in the intro, if not, let's add it.}
%}

% %%%%%%%
% \jw{On the other hand, there are two main weaknesses of connecting entities directly to chunks. First, a chunk is still long enough to contain much information that is beyond a certain entity, which might add additional noise. Second, aggregating multiple chunks can easily add up to hundreds of thousands of tokens, which might hit the long context performance issues such as lost in the middle~\cite{liu2024lost} or low coverage and citation performance~\cite{laban2024summary}.}
% Inspired by recent work on retrieval granularity~\citep{liu2023towards,chen2023dense}, we observe that a factoid or atomic content unit is one of the most basic granularities to represent meaningful information.
% \jw{Thus, we propose to first extract several ``facts'' from each chunk, and} along with entities, they can capture the fine-grained knowledge to build our relatedness tree.
% %%%%%%%%%%
% % For a further synthesis of information, we summarize each cluster, and these summary nodes become the parent of fact aggregates. Same as our similarity tree, recursive summaries are built on upper levels.

\paragraph{Extracting Propositions and Entities from Documents:}
We define a proposition as ``a factual statement describing important information (preferably about some entities) from a paragraph''. We extract entities and propositions using the Distill-SynthKG pipeline \citep{choubey2024distillsynthkgdistillingknowledgegraph}, adapting its SynthKG workflow. 
First, we rewrite chunks of 10K documents from the BAAI/IndustryCorpus\footnote{\url{https://huggingface.co/datasets/BAAI/IndustryCorpus}} to resolve entity references, using \texttt{Meta-Llama-3-70B-Instruct}\footnote{\url{https://huggingface.co/meta-llama/Meta-Llama-3-70B-Instruct}}~\citep{llama3modelcard} with the rewriting prompt shown in Figure~\ref{prompt:decontextualization}. Next, we prompt the same LLM to extract entities from these rewritten chunks (prompt is shown in Figure~\ref{fig:entity-prompt}). After obtaining these entities, we again prompt the LLM to identify all relevant propositions and their associated entities (prompt is shown in Figure~\ref{fig:prop-prompt}). We then consolidate the resulting propositions and entities to fine-tune \texttt{Mistral-7B-Instruct-v0.3}\footnote{\url{https://huggingface.co/mistralai/Mistral-7B-Instruct-v0.3}}~\citep{jiang2023mistral7b}. This smaller fine-tuned model is subsequently used to extract propositions and their associated entities from our multihop datasets.

Our prompt for extracting propositions and entities does not require every proposition to have associated entities. When we prompt LLMs to generate entities for each proposition, they sometimes produce common nouns as entities for those propositions that lack actually associated entities. This can lead to the clustering of unrelated propositions based on common nouns, potentially introducing noise into the relatedness tree.
For example, as illustrated in Figure \ref{fig:entity-less-proposition}, we prefer to avoid LLMs generating \textit{Drug} as an entity for \textit{proposition 1} due to its ambiguity. 
Subsequently, we exclude propositions without associated entities when constructing the relatedness tree, ensuring that only high-quality, entity-linked propositions are utilized. The Table~\ref{tab:facts-stats} provides a detailed breakdown of key statistics for extracted propositions and entities from the MuSiQue, 2Wiki, and HotpotQA datasets. %Notably, key statistics, including the average number of propositions per chunk or entity and the average number of entities per chunk, remain consistent on all three datasets.

\begin{table}[!t]
    \centering
    \caption{Key Statistics for extracted propositions and entities from MuSiQue, 2Wiki, and HotpotQA datasets. We show the number of chunks, propositions, entities, proposition aggregates, and the average, maximum, and minimum number of propositions per entity across the three datasets. 
    % As indicated, indexing approaches adopt either GPT-3.5-Turbo or GPT-4o as the LLM. 
    % Our retrieval baselines do not require the usage of an LLM, and we place them under GPT-4o part.
    }
    \label{tab:facts-stats}
    % \resizebox{\textwidth}{!}{%
    \begin{tabular}{lcccc|ccc}
        \toprule
        & & & & \multicolumn{1}{c}{}& \multicolumn{3}{c}{\#Props per Entity} \\
        % \cmidrule(lr){2-3} \cmidrule(lr){4-5} \cmidrule(lr){6-7} \cmidrule(lr){8-9}
        \cmidrule(lr){6-8}
        Dataset & \#Chunks & \#Props. & \#Ents & \#Pops Aggs. & Avg. & Max. & Min. \\
        \midrule
        MuSiQue & 11,656 & 54,605 & 50,926  & 20,788 & 2.74 & 168 & 1 \\
        2Wiki & 6,119 & 27,697 & 29,490 & 11,108 & 2.49 & 195 & 1 \\
        HotpotQA & 9,221 & 47,153 & 46,856 & 18,278 & 2.66 & 165 & 1 \\ \hline
    \end{tabular}
    % }
\end{table}

\begin{figure}[!t]
    \centering
    \fbox{
        \parbox{0.96\linewidth}{
            % \centering
        \small{{\textit{Proposition 1}}}: \small{{Drug sales are reaching record highs as new therapies are developed and approved.}}
        
        \small{{\textit{Entities}}}: \small{[]}
        
        \small{{\textit{Proposition 2}}}: \small{{Drug sales for Eli Lilly's Mounjaro and Novo Nordisk's semaglutide are reaching record highs as new therapies are developed and approved.}}
        
        \small{{\textit{Entities}}}: \small{[{Eli Lilly, Mounjaro, Novo Nordisk, Semaglutide}]}
        }
    }
    \caption{Examples of propositions with and without associated entities.}
    \label{fig:entity-less-proposition}
\end{figure}

\paragraph{From Entity-specific propositions to Relatedness Tree:}
We concatenate related propositions that share the same entity using exact match to form proposition aggregates.
We ensure that all propositions from the same document are grouped together and maintain their original order. %from document. 
By treating these proposition aggregates as pseudo-documents, we apply the same clustering pipeline in RAPTOR to obtain recursive summaries at levels above them and build the relatedness tree.
Given that most propositions involve multiple entities, each proposition is associated with several entity clusters, offering two key advantages. First, it effectively mimics soft clustering as a single proposition may belong to multiple aggregates. Secondly, when constructing recursive summaries within the RAPTOR framework, the shared propositions across different aggregates result in high embedding similarity, ensuring that these aggregates remain clustered together even at higher levels in the tree.
%
% , which enables us an explicit modeling of relatedness. This relatedness tree forms the right part of \model{} as shown in Figure~\ref{fig:pipeline}. Note that we do not index individual facts after clustering them into fact aggregates for a consideration of efficiency.
%
% In order to find related information within a dataset, we first extract atomic facts and their entities via GPT-4o. \jw{Prafulla/Nan we need to elaboration more here! We need to split this into multiple steps: how we first resolve coreference resolution, then we how we extract entities, how we define "facts", what is the definition and give some examples of that, we also need to give some statistics about facts, how many per chunks we get and how long are they, and how many facts per entities got, from max to min, etc.}
% Each fact is a simple statement that represents a piece of knowledge. Then, we concatenate related facts together to form fact aggregates when a group of facts share the same entity. Since the facts we extracted often have more than one entities, we are essentially conducting soft clustering to group related information. For a further synthesis of information, we apply recursive summaries on levels above fact aggregates by using the same method as RAPTOR, which enables us an explicit modeling of relatedness. 
This relatedness tree forms the right part of \model{} as shown in Figure~\ref{fig:pipeline}. 
Note that we only index aggregated propositions instead of individual propositions for better inference efficiency. 
% a consideration of efficiency.
% 
% \jw{I think we missed the experiments on how different levels of trees are helping the performance, similar to what Raptor did in Table 8 in their paper. Let's see if we can add it, otherwise, we can wait until rebuttal if people asked or the final paper.}

\subsection{Indexing similarity and relatedness trees}
\label{sec:index_both_method}
% As the similarity and relatedness trees have different objectives, w
% e build both trees independently, %. %In other words, we separate each tree during their construction so that summary nodes of a tree do not have access to the clusters of the other tree. 
% % Each tree is constructed independently, 
% ensuring that summary nodes in one tree do not access the clusters of the other. By doing so, we ensure a more efficient tree construction process due to the fewer summaries needed.

% \jw{Prafulla will expand this.}

We propose to construct similarity and relatedness trees independently. This approach ensures that summary nodes in one tree do not access the clusters of the other, leading to a simpler design. There is another slightly complex design in which we allow summary nodes from one tree to access clusters from the other tree. This interaction may enable summary nodes in both trees to inform and enhance each other, improving their informativeness and consequently performance.
However, this approach sacrifices the distinction between similarity and relatedness.
%, with more nodes to be clustered. 
Additionally, allowing cross-tree interaction leads to more nodes to cluster at each level as well as requires summarization based on a greater number of nodes per cluster, all of which increases the overall complexity of the system.
We experimented with both settings and did not observe performance improvement as shown in Appendix~\ref{sec:alternative-cross-tree}. Therefore, we opted for the simpler first implementation in our evaluation.

Flattening all tree nodes, we place them into a unified retrieval pool. In other words, regardless of a node's origin (\emph{e.g.}, bottom or upper levels, similarity or relatedness trees), it is added to a single list containing all nodes.

\section{Experiment Setup}
\subsection{Datasets}
To demonstrate the effectiveness of \model{}, we select three representative multihop QA datasets: MuSiQue~\citep{trivedi2022musiquemultihopquestionssinglehop}, 2WikiMultiHopQA~\citep{ho-etal-2020-constructing}, and HotpotQA~\citep{yang-etal-2018-hotpotqa}. Using the same corpus as HippoRAG~\citep{gutierrez2024hipporag}, we obtain 1000 questions from each validation set of these three datasets.

\subsection{Baselines}
% Two types of baselines are considered:
We select RAPTOR, HippoRAG, and GraphRAG as state-of-the-art retrieval baselines. As discussed above, RAPTOR integrates knowledge based on similarity while the other two approaches focus on relatedness. Specifically, HippoRAG has both indexing and retrieval components, and we use ColBERTv2~\citep{santhanam-etal-2022-colbertv2} as the retriever of HippoRAG due to its strongest QA performance reported. Although GraphRAG has a different goal (global questions directed at an entire dataset) than ours, we include it to show its performance on multihop QA datasets. Since the queries in our datasets ask fine-grained details, we use the local search function of GraphRAG instead of its global search. Additional details of GraphRAG are specified in Appendix~\ref{sec:GraphRAG-details}.

% \item \textbf{Embedding and reranking-based methods} We select BM25~\citep{robertson1994some} and ColBERTv2~\citep{santhanam-etal-2022-colbertv2} as our embedding and reranking-based baselines. Although \model{} is an indexing approach, we can still compare it against these baselines to showcase its strong performance on multihop QA. We also run these baselines on the retrieval pool of \model{} to demonstrate the utility of \model{} on existing retrieval methods.
% \end{itemize}

\subsection{Evaluation metrics}
\label{sec:metrics}
We use exact match (EM) and F1 scores to measure the QA performance of different models. Both metrics evaluate how accurate a generated answer is with respect to the ground truth. Like RAPTOR~\citep{sarthi2024raptor}, we do not assess retrieval performance directly. The reason is that both \model{} and RAPTOR create new candidates (\emph{e.g.}, summary and proposition aggregate) in the retrieval pool, so it would be unfair to compare methods in terms of retrieval scores across different pools. Instead, QA performance is the best indicator of the overall capability of both RAG pipelines.

We use the average time per query (TPQ) and the time-pool efficiency ratio (TPER) to measure the efficiency of \model{} and RAPTOR, as both methods share a significant portion of their retrieval candidates. Average TPQ measures the average time (in seconds) taken to answer a query, and it represents the inference time of a method. For TPER, it computes the growth of total inference time with respect to the growth of the retrieval pool size between two methods:

\begin{equation}
\begin{aligned}
\text{TPER} = \frac{\text{Inference-Time A} / \text{Inference-Time B}}{\text{Pool-Size A} / \text{Pool-Size B}}
\end{aligned}
\label{TPER}
\end{equation}

Setting \model{} as method A and a baseline as method B, we aim to ensure that the growth of inference time does not scale proportionally with the increase in the retrieval pool size. The reason behind is that there are many efficiency considerations (\emph{e.g.}, length and redundancy of retrieval candidates) beyond just the sheer number of retrieval candidates. Parallelization could also be designed to retrieve candidates simultaneously from similarity and relatedness trees, thereby minimizing the effect of retrieval pool size. A TPER value less than 1 indicates reasonable efficiency, whereas a TPER value greater than 1 signifies low efficiency.

% We use average time per query (TPQ) and time per retrieval pool size (TPRS) to measure the efficiency of \model{} and RAPTOR. Average TPQ measures the average time (in seconds) taken to answer a query, and it represents the inference time of a method. For TPRS, it is computed as the total inference time for all queries in a dataset divided by the retrieval pool size of that dataset:

% \begin{equation}
% \begin{aligned}
% \text{TPRS} = \frac{\text{Total Inference Time}}{\text{Size of Retrieval Pool}}\nonumber
% \end{aligned}
% \end{equation}

% In this way, we normalize the total inference time by the size of retrieval pool, which allows us to show the efficiency of a method regardless of its pool size. The reason behind is that there are many efficiency considerations (\emph{e.g.}, length and redundancy of retrieval candidates) beyond just the sheer number of retrieval candidates, so we adopt TPRS to highlight the effect of other factors. A lower score of TPQ or TPRS indicates higher efficiency.

\subsection{Implementation details}
\label{sec:implementation}
To generate final answer, we use GPT-4o and the same prompt (\textit{``answer this question in as fewer number of words as possible.''}) to answer queries for all methods, since we aim to control the instruction-following capabilities across all methods. 
We use either GPT-3.5-Turbo or GPT-4o as the choice of LLM if any methods involve LLM calls.
We use OpenAI's \texttt{text-embedding-3-small} as the embedding model for all methods. During retrieval, we select top 20 candidates that match the provided query for all methods, because there is a large number of text chunks in our datasets and \model{} is expected to perform better when retrieving more due to the incorporation of proposition aggregates and their recursive summaries. 
% Hyperparameters of each model are specified in Appendix~\ref{sec:hyperparameters}.

\section{Results and Analysis}
Our results and analysis aim to answer the following research questions:
\begin{itemize}[noitemsep,topsep=0pt,leftmargin=0.4cm]
    \item \textbf{RQ 1}: How does \model{} compare against other state-of-the-art baselines (sec \ref{sec:overall_results})?
    \item \textbf{RQ 2}: As an important contribution of \model{}, is considering both similarity and relatedness an effective method (sec \ref{sec:overall_results} and \ref{sec:ablation})?
    \item \textbf{RQ 3}: What is the effect of each component in \model{}(sec \ref{sec:ablation})?
    \item \textbf{RQ 4}: What is the applicability of \model{}(sec \ref{sec:applicability})?
    \item \textbf{RQ 5}: With the addition of relatedness tree, is \model{} an efficient method (sec \ref{sec:efficiency})?
\end{itemize}

\begin{table}[ht]
    \centering
    \caption{QA performance of \model{} and baselines. As elaborated in Section~\ref{sec:implementation}, GPT-4o is used to handle QA for all models, and we use two different LLMs (specified in the parentheses) to build indexing structures. We highlight the best scores using either LLM for indexing in green color.
    % As indicated, indexing approaches adopt either GPT-3.5-Turbo or GPT-4o as the LLM. 
    % Our retrieval baselines do not require the usage of an LLM, and we place them under GPT-4o part.
    }
    \label{tab:QA-performance}
    \resizebox{\textwidth}{!}{%
    \begin{tabular}{lcccccc|cc}
        \toprule
        & \multicolumn{2}{c}{MuSiQue} & \multicolumn{2}{c}{2Wiki} & \multicolumn{2}{c}{HotpotQA} & \multicolumn{2}{c}{Average} \\
        \cmidrule(lr){2-3} \cmidrule(lr){4-5} \cmidrule(lr){6-7} \cmidrule(lr){8-9}
        Model & EM & F1 & EM & F1 & EM & F1 & EM & F1 \\
        \midrule
        % \multicolumn{9}{c}{GPT-3.5-Turbo} \\ \midrule
        HippoRAG (GPT-3.5-Turbo) & 32.60 & 43.78 & \cellcolor{green!20}66.40 & \cellcolor{green!20}74.01 & 59.90 & 74.29 & 52.97 & 64.03\\
        RAPTOR (GPT-3.5-Turbo)  & 35.30 & 47.47 & 54.90 & 61.20 & 58.10 & 72.48 & 49.43 & 60.38\\ 
        % \midrule
        % \multicolumn{9}{c}{GPT-4o} \\ \midrule
        GraphRAG (GPT-4o)  & 12.10 & 20.22 & 22.50 & 27.49 & 31.70 & 42.74 & 22.10 & 30.15\\
        % BM25 & 25.90 & 35.88 & 53.00 & 58.58 & 57.70 & 71.32 & 45.53 & 55.26\\
        % ColBERTv2 & 34.00 & 46.80 & 52.90 & 59.48 & 59.00 & 73.42 & 48.63 & 59.90\\
        RAPTOR (GPT-4o)  & 36.40 & 49.09 & 53.80 & 61.45 & 58.00 & 73.08 & 49.40 & 61.21\\ \hline
        \model{} (GPT-3.5-Turbo) & \cellcolor{green!20}38.90 & \cellcolor{green!20}52.08 & 60.40 & 68.20 & \cellcolor{green!20}62.50 & \cellcolor{green!20}77.36 & \cellcolor{green!20}53.93 & \cellcolor{green!20}65.88\\
        \model{} (GPT-4o) &  \cellcolor{green!20}40.50 & \cellcolor{green!20}53.08 & \cellcolor{green!20}59.60 & \cellcolor{green!20}67.94 & \cellcolor{green!20}61.70 & \cellcolor{green!20}76.48 & \cellcolor{green!20}53.93 & \cellcolor{green!20}65.83\\
        \bottomrule
    \end{tabular}
    }
\end{table}

\subsection{Overall results}
\label{sec:overall_results}
Our overall results are presented in Table~\ref{tab:QA-performance}. We show results on more datasets (single-hop QA, other multihop QA, and ambiguous questions) in Appendix~\ref{sec:single-hop} to show the generality of \model{} across various complex reasoning tasks. Besides quantitative scores, we also conduct our qualitative analysis in Appendix~\ref{sec:example-tree}.
% Besides showing the performance on selected benchmarks, we compute the average score of EM and F1.

\textbf{Improvement over baselines} \model{} delivers consistent improvement over RAPTOR, HippoRAG, and GraphRAG. With an exception on 2Wiki when comparing against HippoRAG, \model{} achieves significantly higher performance than indexing baselines (\emph{e.g.}, approximately 5\% higher than RAPTOR on average F1, up to 8.3\% improvement of F1 on MuSiQue than HippoRAG, and more than 20\% higher than GraphRAG on average EM and F1). This demonstrates the advantage of \model{} on multihop QA and modeling both similarity and relatedness. 
Specifically, \model{} outperforms RAPTOR due to the incorporation of a relatedness tree, and it has better overall performance than HippoRAG, because we explicitly model similarity while HippoRAG prioritizes relatedness signals such as nodes with the most edges. 
We see that HippoRAG is particularly strong on 2Wiki benchmark, which is also reported in its original paper. Thus, we believe 2Wiki is the best fit of HippoRAG, but it has lower performance scores than \model{} on other datasets. One potential reason is that the entity-centric design of 2Wiki may be well-suited for HippoRAG, as noted in the HippoRAG paper.
% \jw{Any characteristic of 2wiki that is causing this issue? larger number of candidates does not explain why it only works on 2wiki but not others.}

As for GraphRAG, it considers relatedness solely, and it delivers the worst performance scores on our datasets. After a manual verification, we find that GraphRAG often provides ``I don't know'' answers, suggesting that it prefers not to give a concrete answer. Since GraphRAG is designed to handle query-focused summarization of an entire corpus, it is not the most competitive approach in terms of accuracy for existing multihop QA tasks.

% \textbf{Improvement over retrieval baselines} \model{} offers significant improvement over retrieval baselines as well. For example, based on average scores, it shows more than 5\% improvement over ColBERTv2, and ColBERTv2 is a stronger retrieval method than BM25. This shows that the great potential of performing indexing before retrieval. Since \model{} simply uses the embeddings generated from \texttt{text-embedding-3-small} to retrieve candidates, we also think that this embedding model contributes to strong retrieval performance.

\textbf{Effect of LLM choice} When comparing the performance of \model{} using GPT-4o as the LLM against itself using GPT-3.5-Turbo, we find the QA performance is not significantly affected. This phenomenon also holds on RAPTOR. Since the choice of LLM for \model{} and RAPTOR only affects summarization results, we believe GPT-3.5-Turbo is a sufficiently good option for both methods. This allows researchers to pursue a more cost-effective solution with \model{} for indexing.
% Therefore, this is an advantage of \model{} and RAPTOR, because their performance on multihop QA is not dependant on a much stronger LLM. This advantage allows researchers to pursue a more cost-effective solution when using \model{} to perform indexing.

\begin{table}[t]
    \centering
    \caption{Ablation study of \model{}. 
    % Starting from RAPTOR tree (similarity tree), we iteratively add a component to it. We first add propositions (\textbf{second row}) or proposition aggregates (\textbf{third row}). Then, on top of RAPTOR+proposition, we add recursive summaries (\textbf{fourth row}). We also study the performance of adding proposition aggregates to RAPTOR+Proposition+Summary (\textbf{fifth row}), so the resulting indexing tree becomes the same as Figure~\ref{fig:pipeline}, except for changing the dashed lines to solid. Finally, we study the combination of proposition aggregates and text chunks in the same pool for finding clusters and performing summarization (\textbf{sixth row}), which unifies the similarity and relatedness trees instead of keeping them separate.
    }
    \label{tab:ablation}
    \resizebox{\textwidth}{!}{%
    \begin{tabular}{lcccccc|cc}
        \toprule
        & \multicolumn{2}{c}{MuSiQue} & \multicolumn{2}{c}{2Wiki} & \multicolumn{2}{c}{HotpotQA} & \multicolumn{2}{c}{Average} \\
        \cmidrule(lr){2-3} \cmidrule(lr){4-5} \cmidrule(lr){6-7} \cmidrule(lr){8-9}
        Variants & EM & F1 & EM & F1 & EM & F1 & EM & F1 \\
        \midrule
        \model{}  &  \cellcolor{green!20}40.50 & \cellcolor{green!20}53.08 & \cellcolor{green!20}59.60 & \cellcolor{green!20}67.94 & \cellcolor{green!20}61.70 & \cellcolor{green!20}76.48 & \cellcolor{green!20}53.93 & \cellcolor{green!20}65.83\\
        (A): \model{} $-$ Re. Summary & 37.50 & 50.38 & 57.70 & 65.75 & 61.20 & 75.99 & 52.13 & 64.04\\
        % (B): (A) $-$ Aggregate  & 36.40 & 49.09 & 53.80 & 61.45 & 58.00 & 73.08 & 49.40 & 61.21\\ --> Raptor
        (B): \model{} $+$ Proposition & 39.10 & 51.80 & 58.10 & 65.53 & 60.80 & 75.26 & 52.67 & 64.20\\
        (C): (B) $-$ Aggregate & 34.70 & 47.82 & 53.20 & 60.22 & 58.90 & 73.38 & 48.93 & 60.47\\
        (D): (C) $-$ Re. Summary  & 33.90 & 46.19 & 53.00 & 59.13 & 57.00 & 71.09 & 47.97 & 58.80\\
        (E): Dual Clustering on Chunks & 34.80 & 47.32 & 53.50 & 59.93 & 56.60 & 71.84 & 48.30 & 59.70\\
        % \model{} (w/o separation) & 39.70 & 52.63 & 59.90 & 67.93 & 62.80 & 77.64 & 54.13 & 66.07\\
        
        \bottomrule
    \end{tabular}
    }
\end{table}

\subsection{Ablation study}
\label{sec:ablation}
To dissect \model{}, we perform a comprehensive ablation analysis as shown in Table~\ref{tab:ablation}. 
There are several variances, including 
(A) remove the recursive summary on the relatedness tree;
(B) add all the propositions into the retrieval pool, and keep all aggregated propositions and recursive summary on the relatedness tree;
(C) same as (B) but remove aggregated propositions;
(D) same as (C) but further remove the recursive summary design on the relatedness tree.

\textbf{Entity clustering} In (E), we do not maintain a separate relatedness tree and add an additional clustering philosophy to the similarity tree. Specifically, each text chunk in the similarity tree is simplified to, 'This chunk mentions \textit{entity 1} and \textit{entity 2},' if both entities are extracted by our LLM. We then run GMMs (the same clustering method as RAPTOR) on these simplified chunks. Once the clustering decisions are obtained, we group the original chunks as additional clusters and append these clusters to the similarity tree, allowing higher levels of the tree to incorporate both clustering philosophies. Since entities primarily determine the outcome of this additional clustering approach, we apply entity clustering to model relatedness on the similarity tree. This allows us to eliminate proposition aggregates in order to examine their utility.

\textbf{Findings} Overall, we observe performance drops across all variations, highlighting the effectiveness of our design for \model{}. First, the recursive summary on the relatedness tree proves beneficial, as seen in both (A) and (D). Interestingly, adding more propositions to retrieval negatively impacts performance, as shown in (B). This indicates adding redundant information into the retrieval pool hurts the QA performance, since we keep the aggregated propositions in \model{}. From (C), we also find that aggregated propositions are essential, with their removal resulting in a significant performance decline. This is an important indicator that adding grouped knowledge about relatedness to the similarity tree would offer improvements, which echoes the bottlenecks described in Section~\ref{sec:bottleneck}.

Both (A) and (B) have better performance than RAPTOR (GPT-4o) from Table~\ref{tab:QA-performance}, which indicates the advantage of proposition aggregates. In contrast, although (E) also models both similarity and relatedness, it exhibits a notable decline comparing against \model{}. This finding demonstrates the necessity of proposition aggregates of modeling relatedness. Because proposition aggregates reduce noise and information redundancy more effectively than text chunks as described in Section~\ref{sec:relatedness_method}, they serve as an effective carrier of related dataset contents.

\begin{table}[t]
    \centering
    \caption{Applicability of \model{} when a specific retrieval method is selected. We feed our non-indexing models with the retrieval pool of \model{} and see whether QA performance improves.}
    \label{tab:applicability}
    \begin{tabular}{lcccccc|cc}
        \toprule
        & \multicolumn{2}{c}{MuSiQue} & \multicolumn{2}{c}{2Wiki} & \multicolumn{2}{c}{HotpotQA} & \multicolumn{2}{c}{Average} \\
        \cmidrule(lr){2-3} \cmidrule(lr){4-5} \cmidrule(lr){6-7} \cmidrule(lr){8-9}
        Variants & EM & F1 & EM & F1 & EM & F1 & EM & F1 \\
        \midrule
        BM25 & 25.90 & 35.88 & 53.00 & 58.58 & 57.70 & 71.32 & 45.53 & 55.26\\
        RAPTOR + BM25 & 27.00 & 38.91 & 50.60 & 57.00 & 56.90 & 70.88 & 44.83 & 55.60\\
        \model{} + BM25 & \cellcolor{green!20}35.00 & \cellcolor{green!20}47.66 & \cellcolor{green!20}58.20 & \cellcolor{green!20}65.72 & \cellcolor{green!20}61.70 & \cellcolor{green!20}75.88 & \cellcolor{green!20}51.63 & \cellcolor{green!20}63.09\\
        ColBERTv2 & 34.00 & 46.80 & 52.90 & 59.48 & 59.00 & 73.42 & 48.63 & 59.90\\
        RAPTOR + ColBERTv2 & 34.20 & 47.17 & 50.30 & 57.51 & 57.90 & 72.31 & 47.47 & 59.00\\
        \model{} + ColBERTv2 & \cellcolor{green!20}38.10 & \cellcolor{green!20}51.32 & \cellcolor{green!20}56.70 & \cellcolor{green!20}64.74 & \cellcolor{green!20}60.90 & \cellcolor{green!20}75.72 & \cellcolor{green!20}51.90 & \cellcolor{green!20}63.93 \\
        \bottomrule
    \end{tabular}
\end{table}

\subsection{Applicability of \model{}}
\label{sec:applicability}
We analyze how applicable \model{} is when a specific retrieval method is chosen. Therefore, we select BM25~\citep{robertson1994some} and ColBERTv2~\citep{santhanam-etal-2022-colbertv2} as additional reranking-based options. We run them on the retrieval pool of \model{} to demonstrate its utility. 
We show how \model{} can complement these non-indexing options in Table~\ref{tab:applicability}. 
Results show that having \model{} benefits both BM25 and ColBERTv2 significantly, which demonstrates the advantage of our solution as the upstream step of these methods. 
On the other hand, RAPTOR only improves the QA performance on MiSuQue while showing performance degradation on other datasets. Thus, the utility of \model{} surpasses that of RAPTOR in the context of multihop reasoning. We also apply \model{} on an iterative retrieval method called self-ask~\citep{press-etal-2023-measuring} and obtain significant performance improvement as shown in Appendix~\ref{sec:self-ask}. Our method showcases wide applicability on multihop QA across various retrieval methods.

\subsection{Efficiency of \model{}}
\label{sec:efficiency}
As shown in Table~\ref{tab:efficiency}, we compare the efficiency of \model{} and RAPTOR using the metrics described in Section~\ref{sec:metrics}. All the values listed involve the time taken to retrieve the top 20 candidates and prompt GPT-4o to answer the query.
% Besides the strong QA performance of \model{}, efficiency is also an important factor to analyze. We find that GraphRAG
% % both HippoRAG \jw{TODO} and GraphRAG 
% is not comparable to tree-based methods in terms of inference time and the total time required to construct indexing, as its knowledge graphs tend to be more complex than simple trees as in \model{}. Therefore, as shown in Table~\ref{tab:efficiency}, we compare the efficiency of \model{} and RAPTOR using the metrics described in Section~\ref{sec:metrics}. All the inference times listed in the table involve the time taken to retrieve the top 20 candidates and prompt GPT-4o to answer the query.

RAPTOR requires less inference time than \model{} on average, which is expected due to \model{}'s larger retrieval pool. However, with slightly longer inference time, \model{} has much better performance as discussed previously. To evaluate whether \model{} remains a reasonably efficient method, we compute its TPER values to measure its growth of total inference time relative to its growth of retrieval pool size. Since all its TPER values are well below 1, \model{} demonstrates reasonable efficiency without introducing many lengthy or redundant retrieval candidates.

% we compute TPRS that normalizes the total inference time by the size of retrieval pool each method has. It is clear to see that \model{} always has lower TPRS than RAPTOR, which indicates a low level of information redundancy. This also shows that \model{} is an efficient method given its retrieval pool size.

\begin{table}[t]
    \centering
    \caption{Efficiency of \model{} and RAPTOR. Since both methods share a significant portion of retrieval candidates, we designate \model{} as Method A and RAPTOR as Method B in the TPER columns, as defined in Equation ~\ref{TPER}.}
    \label{tab:efficiency}
    \begin{tabular}{lcccccc|cc}
        \toprule
        & \multicolumn{2}{c}{MuSiQue} & \multicolumn{2}{c}{2Wiki} & \multicolumn{2}{c}{HotpotQA} & \multicolumn{2}{c}{Average} \\
        \cmidrule(lr){2-3} \cmidrule(lr){4-5} \cmidrule(lr){6-7} \cmidrule(lr){8-9}
        Model & TPQ$\downarrow$ & TPER$\downarrow$ & TPQ$\downarrow$ & TPER$\downarrow$ & TPQ$\downarrow$ & TPER$\downarrow$ & TPQ$\downarrow$ & TPER$\downarrow$ \\
        \midrule
        RAPTOR  & 1.560 & - & 1.437 & - & 1.502 & - & 1.500 & -\\
        \model{}  & 2.653 & 0.600 & 1.974 & 0.499 & 2.319 & 0.517 & 2.315 & 0.539 \\
        \bottomrule
    \end{tabular}
\end{table}

\section{Conclusion}
In this paper, we identify the bottleneck of solely modeling similarity or relatedness when we need to index a multihop reasoning dataset for knowledge integration. To address it, we introduce \model{}, an innovative RAG indexing approach that considers both similarity and relatedness. \model{} delivers a consistent improvement over state-of-the-art indexing baselines across several multihop QA benchmarks.

\bibliography{iclr2025_conference}
\bibliographystyle{iclr2025_conference}

\appendix
% \section{Hyperparameters}
% \label{sec:hyperparameters}
% You may include other additional sections here.
\section{An Alternative Design of Allowing Cross-Tree Interaction}
\label{sec:alternative-cross-tree}
We discuss an alternative design of combining similarity and relatedness trees. Specifically, this design combines nodes from both sides in the same pool for finding additional clusters and performing summarization at every tree level. In other words, we find additional clusters by concatenating the nodes of both trees, which considers cross-tree interaction instead of keeping them separate. 

As shown in Table~\ref{tab:cross-tree}, the performance of considering cross-tree interaction is slightly lower than \model{}. Therefore, it is more efficient to keep trees separate in order to reduce the overall complexity of the system as discussed in Section~\ref{sec:index_both_method}.

\section{Additional Experiment on Other Non-Indexing Methods}
\label{sec:self-ask}
Although there are many existing methods that work on multihop reasoning tasks, \model{} is about indexing corpus data under RAG setup. In other words, instead of being our baselines, other non-indexing works\citep{press-etal-2023-measuring,islam-etal-2024-open} focus on other dimensions of improving performance on complex reasoning tasks.

To further demonstrate the applicability of \model{}, we have run two additional sets of experiments: (1) the closed-book setting, and (2) an iterative retrieval method~\citep{press-etal-2023-measuring} called self-ask specifically designed for multihop reasoning. The closed-book setting means to directly get the final answer from GPT-4o without any retrieval. For self-ask, we prompt GPT-4o in two iterations without using a search engine. In the first iteration, the model is prompted to propose follow-up questions and provide answers to them. In the second iteration, GPT-4o is instructed to answer the final question by incorporating the follow-up thought process. We feed the model with a one-shot example and 10 retrieved candidates that match the final question in both iterations. For the self-ask retrieval pool, we use either all text chunks or \model{}.

Similar to Table~\ref{tab:applicability}, our experiment in Table~\ref{tab:self-ask} on self-ask shows that \model{} can complement existing methods for optimal performance. We see that the closed-book setting yields the worst performance, which indicates that LLMs’ parametric knowledge alone does not offer a decent performance on our datasets. Then, we see \model{} successfully improves the scores of self-ask, demonstrating its wide applicability. By leveraging \model{}'s retrieval pool, we view our method as an augmentation to other non-indexing methods for multihop reasoning, rather than as a competitor.

Efficiency-wise, we show TPQ of using both \model{} and self-ask in Table~\ref{tab:efficiency_self-ask}. By having \model{}, the TPQ of self-ask increases by approximately $1.2$ seconds over the three datasets. Since self-ask requires two iterations of LLM prompting in our implementation, the increase in TPQ is relatively small compared to the significant performance improvement brought by \model{}.

\section{Retrieval Pool Size}
The retrieval pool sizes of \model{} on MuSiQue, 2Wiki, and HotpotQA are 35070, 19100, and 29934 respectively. The retrieval pool sizes of RAPTOR on MuSiQue, 2Wiki, and HotpotQA are 12371, 6939, and 10031 respectively. \model{}'s retrieval pool size is slightly less than three times the size of RAPTOR's. Considering the discussion on TPER in Section~\ref{sec:efficiency}, we believe our method is reasonably efficient.

\section{An Example \model{} Tree}
\label{sec:example-tree}
\begin{figure}[h]
    \centering
    % \vspace{0.3cm}
    \includegraphics[width=\textwidth]{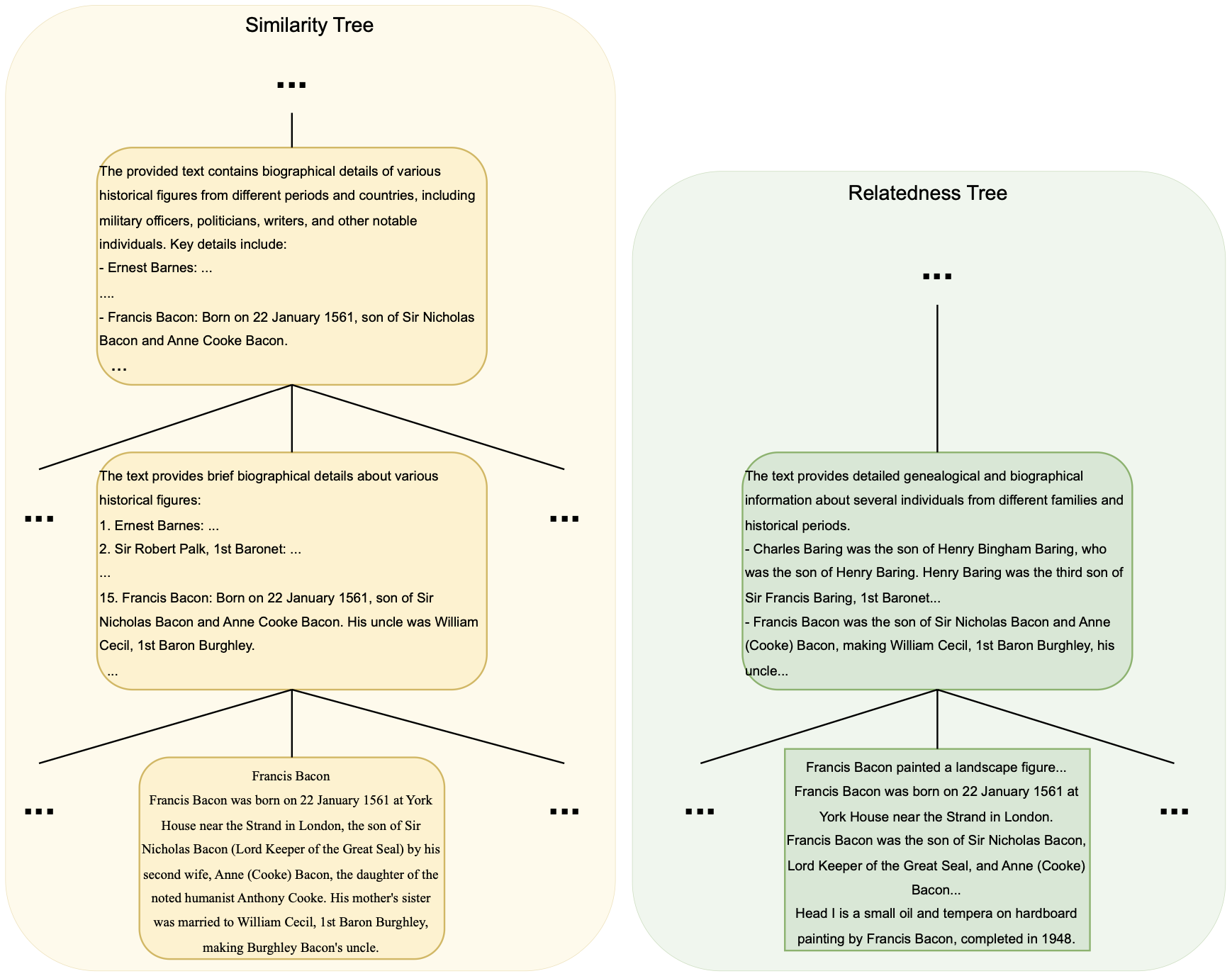}
    \caption{Relevant part of the \model{} tree for the question: ``who is the father of the artist who painted Head I?''.}
    \label{fig:example-tree}
    % \vspace{-0.3cm}
\end{figure}

Using the question ``who is the father of the artist who painted Head I?'' as an example, we focus on the relevant part of the \model{} tree in Figure~\ref{fig:example-tree} to conduct our qualitative analysis.

\begin{table}[t]
    \centering
    \caption{QA performance of two designs: separating similarity and relatedness trees (\model{}) and allowing cross-tree interaction.}
    \label{tab:cross-tree}
    \resizebox{\textwidth}{!}{%
    \begin{tabular}{lcccccc|cc}
        \toprule
        & \multicolumn{2}{c}{MuSiQue} & \multicolumn{2}{c}{2Wiki} & \multicolumn{2}{c}{HotpotQA} & \multicolumn{2}{c}{Average} \\
        \cmidrule(lr){2-3} \cmidrule(lr){4-5} \cmidrule(lr){6-7} \cmidrule(lr){8-9}
        Variants & EM & F1 & EM & F1 & EM & F1 & EM & F1 \\
        \midrule
        \model{}  &  \cellcolor{green!20}40.50 & \cellcolor{green!20}53.08 & \cellcolor{green!20}59.60 & \cellcolor{green!20}67.94 & \cellcolor{green!20}61.70 & \cellcolor{green!20}76.48 & \cellcolor{green!20}53.93 & \cellcolor{green!20}65.83\\
        Cross-tree interaction & 40.20 & 53.06 & 58.30 & 65.25 & 60.30 & 75.71 & 52.93 & 64.67\\
        \bottomrule
    \end{tabular}
    }
\end{table}

\begin{table}[t]
    \centering
    \caption{QA performance of closed-book and self-ask. We feed self-ask with the unified retrieval pool of \model{} and see whether performance benifits from that.}
    \label{tab:self-ask}
    \resizebox{\textwidth}{!}{%
    \begin{tabular}{lcccccc|cc}
        \toprule
        & \multicolumn{2}{c}{MuSiQue} & \multicolumn{2}{c}{2Wiki} & \multicolumn{2}{c}{HotpotQA} & \multicolumn{2}{c}{Average} \\
        \cmidrule(lr){2-3} \cmidrule(lr){4-5} \cmidrule(lr){6-7} \cmidrule(lr){8-9}
        Model & EM & F1 & EM & F1 & EM & F1 & EM & F1 \\
        \midrule
        Closed-book & 10.0 & 22.0 & 19.0 & 34.0 & 29.0 & 44.0 & 19.3 & 33.3\\
        Self-ask & 31.20 & 44.35 & 55.00 & 61.99 & 57.10 & 71.11 & 47.77 & 59.15\\
        \model{} + self-ask &  \cellcolor{green!20}36.50 & \cellcolor{green!20}49.12 & \cellcolor{green!20}57.20 & \cellcolor{green!20}65.13 & \cellcolor{green!20}59.70 & \cellcolor{green!20}74.07 & \cellcolor{green!20}51.13 & \cellcolor{green!20}62.77\\
        \bottomrule
    \end{tabular}
    }
\end{table}

\begin{table}[]
\centering
\caption{Efficiency of using self-ask with \model{}. We show TPQ on MuSiQue, 2Wiki, and HotpotQA datasets.}
\label{tab:efficiency_self-ask}
\begin{tabular}{@{}lcccc@{}}
\toprule
Model & MuSiQue TPQ & 2Wiki TPQ & HotpotQA TPQ & Average TPQ \\ \midrule
Self-ask & 2.72 & 2.21 & 2.29 & 2.41 \\
\model{} + self-ask & 4.53 & 3.07 & 3.30 & 3.63 \\ \bottomrule
\end{tabular}
\end{table}

For the multihop question in Figure~\ref{fig:motivation} (correct answer: Nicholas Bacon), the MuSiQue corpus contains one relevant paragraph stating, ``Francis Bacon was born on 22 January 1561 at York House near the Strand in London, the son of Sir Nicholas Bacon...''. This paragraph is a leaf node of the similarity tree shown in Figure~\ref{fig:example-tree}. The similarity tree for the entire MuSiQue corpus has two more mentions (both mentions are in summary nodes) of Nicholas Bacon, one of which reads: ``...Francis Bacon: Born on 22 January 1561, son of Sir Nicholas Bacon and Anne Cooke Bacon...'' The addition of our relatedness tree adds two more mentions of Nicholas Bacon: one is in a proposition aggregate (``...Francis Bacon was the son of Sir Nicholas Bacon, Lord Keeper of the Great Seal, and Anne (Cooke) Bacon... Head I is a small oil and tempera on hardboard painting by Francis Bacon, completed in 1948...''), and the other one is a summary node (''The text provides detailed genealogical and biographical information about several individuals from different families and historical periods... Francis Bacon was the son of Sir Nicholas Bacon and Anne (Cooke) Bacon, making William Cecil, 1st Baron Burghley, his uncle...''). Thus, our similarity tree (RAPTOR tree) has three mentions of Nicholas Bacon, but none of them contains Head I information. \model{} has five mentions of Nicholas Bacon, and one of them (the proposition aggregate) contains Head I information. This proposition aggregate groups several propositions together via the entity ``Francis Bacon''. Because this node is the only retrieval candidate that fully matches the question, retrieving it would maximize the chance of generating the correct answer. If we use the RAPTOR tree only, we will not have this retrieval candidate. We believe this is an excellent example of how a comprehensive knowledge integration process can enhance the performance of RAG in multihop reasoning.

\section{Performance on Single-Hop QA, MultiHop-RAG, and Ambiguous Questions}
\label{sec:single-hop}
To showcase the generality of \model{} on more datasets of complex reasoning tasks, we run the comparison between \model{} and RAPTOR on single-hop questions and MultiHop-RAG dataset~\citep{tang2024multihopragbenchmarkingretrievalaugmentedgeneration}. We also try ASQA dataset~\citep{stelmakh-etal-2022-asqa} that contains ambiguous factoid questions. For single-hop questions, we use MuSiQue dataset and collect all the decomposed questions of multihop queries. We filter out some decomposed questions if they are still multihop or are based on another question. As a result, we end up with 502 single-hop questions from MuSiQue. As for MultiHop-RAG, it is a more recent dataset. We filter all unanswerable questions and randomly select 350 ``comparison'' queries and 350 ``inference'' queries, which forms a pool of 700 queries in total.

Moreover, the primary difference between the multihop QA datasets and ASQA is that ASQA requires LLMs to reason across multiple perspectives (\emph{e.g.}, disambiguated questions) of an ambiguous question and organize their generation into a coherent and detailed answer. We report scores on ASQA based on all 948 ambiguous questions of its development set.

Table~\ref{tab:single-hop} shows the performance scores of RAPTOR and our method. On the single-hop questions, \model{} still outperforms RAPTOR, but the lead narrows compared to the scores in Table~\ref{tab:QA-performance}. Since all queries in MuSiQue involve at least two hops, we observe that an increased number of reasoning hops positively impacts \model{}'s performance. This is because single-hop questions may not require comprehensive knowledge synthesis, as they only involve retrieving the relevant chunks for the single hop. However, with more hops, we not only need to retrieve relevant chunks but also synthesize them comprehensively. \model{} also delivers better performance on MultiHop-RAG, which echoes our main experiment.

Table~\ref{tab:single-hop} also displays STR-EM (string exact match), Disambig-F1, and Disambiguation-Rouge metrics for ASQA dataset. Specifically, STR-EM and Disambig-F1 dissect RAG answers to ambiguous questions into multiple perspectives and measure their accuracy with respect to each disambiguated question. Disambiguation-Rouge serves as an overall statistic that incorporates both ROUGE (with respect to the references) and accuracy scores. We see \model{} delivers a consistent improvement over RAPTOR, which again demonstrates the benefit of adopting \model{} on ambiguous queries.

Our comprehensive selection of datasets demonstrates the generality and contribution of \model{} across various complex reasoning tasks, with the most significant improvement observed in multihop QA.

\section{Additional GraphRAG Details}
\label{sec:GraphRAG-details}
We adhere to GraphRAG documentation\footnote{\url{https://microsoft.github.io/graphrag/}} to construct its indexing structure and handle QA. Following the recommendation, we use its ``auto tuning'' to generate domain adapted prompts for the creation of its knowledge graph. We use the default values for most hyperparameters, except that we set the response type to ``a few words'' and do not include covariates during indexing.

\section{LLM Prompts}
\label{sec:llm-prompts}
\begin{figure}[h]
    \centering
    % \vspace{0.3cm}
    \includegraphics[width=\textwidth]{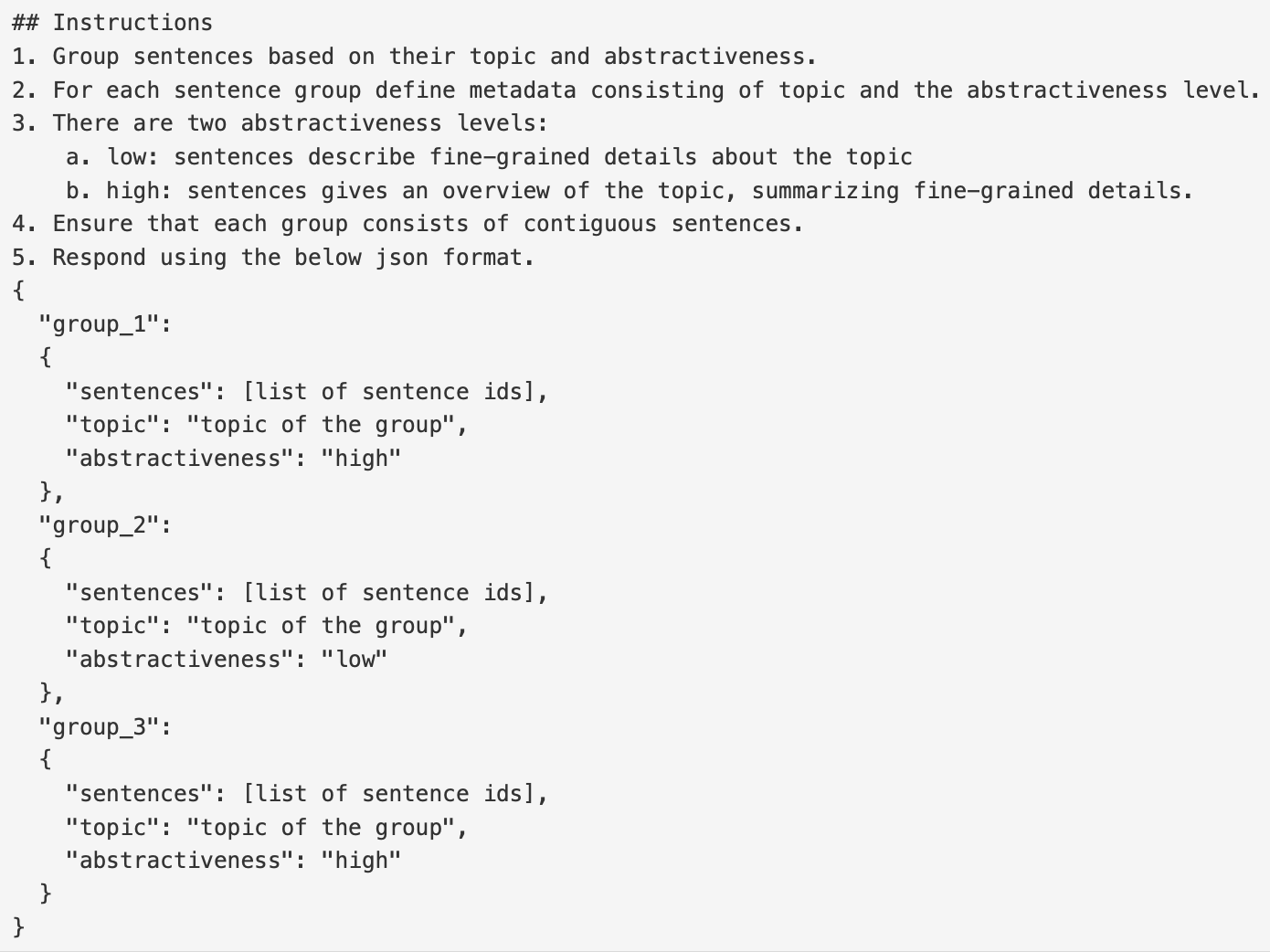}
    \caption{Prompt of identifying a two-level hierarchy for all candidate chunks.}
    \label{fig:hierarchy-prompt}
    % \vspace{-0.3cm}
\end{figure}

The prompt we use to perform summarization on a cluster of nodes is ``summarize the provided text, including as many key details as needed''. This prompt is the same as RAPTOR. In Section~\ref{sec:bottleneck}, we use `identify the high-level topic of this paragraph as concise as possible'' to extract the topic of each passage. As mentioned in Section~\ref{sec:hierarchy}, the prompt used for identifying a two-level hierarchy for all chunks is shown in Figure~\ref{fig:hierarchy-prompt}. As mentioned in Section~\ref{sec:relatedness_method}, the LLM prompt for rewriting chunks is shown in Figure~\ref{prompt:decontextualization}, the prompt for extracting named entities from rewritten chunks is shown in Figure~\ref{fig:entity-prompt}, and the prompt for extracting propositions is shown in Figure~\ref{fig:prop-prompt}.

\begin{table}[t]
    \centering
    \caption{QA performance on single-hop questions, MultiHop-RAG, and ASQA.}
    \label{tab:single-hop}
    \resizebox{\textwidth}{!}{%
    \begin{tabular}{lccccccc}
        \toprule
        & \multicolumn{2}{c}{Single-Hop} & \multicolumn{2}{c}{MultiHop-RAG} & \multicolumn{3}{c}{ASQA} \\
        \cmidrule(lr){2-3} \cmidrule(lr){4-5} \cmidrule(lr){6-8} 
        Model & EM & F1 & EM & F1 & STR-EM & Disambig-F1 & Disambiguation-Rouge \\
        \midrule
        RAPTOR & 73.90 & 80.67 & 85.43 & 86.37 & 51.47 & 40.30 & 42.72\\
        \model{}  &  \cellcolor{green!20}74.10 & \cellcolor{green!20}82.64 & \cellcolor{green!20}87.57 & \cellcolor{green!20}88.41 & \cellcolor{green!20}52.81 & \cellcolor{green!20}41.82 & \cellcolor{green!20}43.47  \\
        \bottomrule
    \end{tabular}
    }
\end{table}

\begin{figure*}[!h]
\small
\begin{mdframed}
\textbf{Previous paragraph from Document}:\\
Gualala, the isolated Mendocino Coast town with a name that leaves most visitors tongue-tied, is on a new list of the 50 best places to live in the United States. Men's Journal magazine describes Gualala as an \"outpost of adventure lifestyle\" in its latest edition, which goes on sale today. The magazine describes Gualala (pronounced wa-LA-la by locals) as one of the \"below-the-radar places to a make a move on before the word gets out.\" There were five such cities. The others were Homer, Alaska; Newport, Vt.; Logan, Utah; and Walla Walla, Wash. Rolling Stone magazine's Jann Wenner publishes Men's Journal, which has a paid circulation of about 620,000. Gualala joined three other California communities on the magazine's list: Santa Cruz, Mammoth Lakes and Bishop. \"We were looking for places that combined affordability, proximity to outdoor adventure and a generally undiscovered quality of life,\" said Erica Kestenbaum, a spokeswoman for Men's Journal.\\
\textbf{Instruction}: \\
Rewrite the below paragraph by resolving all entity coreferences with the preceding paragraph from document. \\
- Resolve all inter-sentence pronoun references. \\
- Make sure that all pronouns in a sentence refers to some named entity with in the same sentence. \\
- Explicitly mention entity names wherever necessary to remove ambiguity from a sentence. Remember to make each sentence clear and unambiguous. \\
- For each entity, use only the one most informative name.\\
- Do not generate anything except the rewritten paragraph. \\
\textbf{Paragraph}: \\
She said isolation played a factor. \"In Northern California, it's particularly difficult to find a beautiful coastal setting that isn't entirely overrun,\" she said. Gualala residents Monday were largely unaware of the magazine listing or the attention it could bring to the old logging town turned tourist center. A few coastal residents chuckled about any notion of affordability, given an influx of newcomers who've driven the median housing price to \$580,000 compared to the median family income of \$47,778. Others recalled an era when the Gualala region was better known for the logging of ancient redwoods, marijuana growing and boisterous beer drinking at the historic Gualala Hotel. Still there was a certain pride to the magazine's designation. Yvette White, a 25-year resident who works at the Gualala Sport; Tackle shop, said she's proud her town made it on the list.\\
\textbf{Output}:\\
Erica Kestenbaum said isolation played a factor. \"In Northern California, it's particularly difficult to find a beautiful coastal setting that isn't entirely overrun,\" Erica Kestenbaum said. Gualala residents Monday were largely unaware of the Men's Journal magazine listing or the attention it could bring to the old logging town turned tourist center. A few coastal residents of Gualala chuckled about any notion of affordability, given an influx of newcomers who've driven the Gualala’s median housing price to \$580,000 compared to the median family income of \$47,778. Other Gualala residents recalled an era when the Gualala region was better known for the logging of ancient redwoods, marijuana growing and boisterous beer drinking at the historic Gualala Hotel. Still there was a certain pride to the Men's Journal magazine's designation. Yvette White, a 25-year Gualala resident who works at the Gualala Sport; Tackle shop, said she's proud her town made it on the list.\\
\textbf{Previous paragraph from Document}: [previous paragraph]\\
\textbf{Instruction}: \\
Rewrite the below paragraph by resolving all entity coreferences with the preceding paragraph from document. \\
- Resolve all inter-sentence pronoun references. \\
- Make sure that all pronouns in a sentence refers to some named entity with in the same sentence. \\
- Explicitly mention entity names wherever necessary to remove ambiguity from a sentence. Remember to make each sentence clear and unambiguous. \\
- For each entity, use only the one most informative name.\\
- Do not generate anything except the rewritten paragraph. \\
\textbf{Paragraph}: [paragraph ]\\
\textbf{Output}:\\
\end{mdframed}
\caption{Prompt for rewriting a paragraph (\emph{e.g.}, a document chunk) by resolving entity coreferences.} \label{prompt:decontextualization}
\end{figure*}

\begin{figure*}[!h]
\tiny
\begin{mdframed}
Extract all named entities from the document. Also generate the type for each entity.

\textbf{Instructions}

- Generate only the most informative name for each named entity. Example: if John P., Parker, John Parker are coreferential, only generate John Parker.\\
- Use your best understanding best on the domain of paragraph to decide appropriate entity types.\\
- Respond using json format provided below.

\begin{verbatim}
{
    "n1":{"name": "entity_name", "type": "entity_type_label"},
    "n2":{},
}
\end{verbatim}

Below is an example for reference.\\
Paragraph: Tucked into Eli Lilly’s year-end earnings report, the company revealed positive results from Synergy-NASH—its phase 2 study of tirzepatide in adults in nonalcoholic steatohepatitis (NASH), also known as metabolic dysfunction-associated steatohepatitis (MASH).\\
Output:

\begin{verbatim}
{
    "n1": {"name": "Eli Lilly", "type": "Organization"},
    "n2": {"name": "Synergy-NASH", "type": "Clinical Trial"},
    "n4": {"name": "tirzepatide", "type": "Drug"},
    "n5": {"name": "nonalcoholic steatohepatitis", "type": "Disease"},
    "n6": {"name": "metabolic dysfunction-associated steatohepatitis", "type": "Disease"},
    "n7": {"name": "year-end earnings report", "type": "Document"}
}
\end{verbatim}
\end{mdframed}
\caption{Prompt for extracting entities from a document (\emph{e.g.}, a rewritten chunk).}
\label{fig:entity-prompt}
\end{figure*}

\begin{figure*}[!h]
\tiny
\begin{mdframed}
Extract all facts from the document. For each fact, also generate all semantic triplets. \\
\textbf{Instructions}\\
- Consistently use the most informative name for each named entity in all facts and triplets. \\
- Avoid pronouns or ambiguous references in facts and triplets. Instead, directly include all relevant named entities in facts. \\
- Ensure that each semantic triplet contains head entity, predicate, and tail entity. \\
- Ensure that at least one (preferably both) entity in each semantic triplet is present in the given entities list. \\
- Respond using json format provided below:\\
\begin{verbatim}
{
    "f1":{
        "fact": "A factual statement describing important information (preferably about some entities) from the paragraph",
        "triplets: [["entity 1", "predicate", "entity 2"], ["entity 1", "predicate", "entity 3"]]
    },
    "f2":{},
}   
\end{verbatim}

Below is an example for reference. \\
Paragraph: Locked in a heated battle with Novo Nordisk’s semaglutide franchise, Eli Lilly’s tirzepatide is beginning to come into its own—both with regards to sales and amid attempts to show the dual GIP/GLP-1 agonist can strike out beyond diabetes and obesity. As Mounjaro, tirzepatide won its first FDA nod in Type 2 diabetes back in May 2022. An obesity approval followed last November, with that formulation of tirzepatide adopting the commercial moniker Zepbound. In 2023’s fourth quarter, Mounjaro generated a whopping \$2.2 billion in sales, a nearly eight-fold increase over the \$279 million it pulled down during the same stretch in 2022. Year-to-date, the drug brought home around \$5.2 billion in revenues, Lilly said in an earnings release Tuesday. Zepbound, for its part, generated \$175.8 million during its first quarter on the market. Overall, Lilly reeled in around \$9.4 billion in fourth-quarter sales, growing 28\% over the \$7.3 billion it made for the quarter in 2022.\\
Entities: Eli Lilly, Novo Nordisk, Tirzepatide, Semaglutide, GLP-1, GIP, FDA, Mounjaro, Zepbound \\
Output:\\
\begin{verbatim}
{
    "f1": {
        "fact": "Eli Lilly's tirzepatide is competing with Novo Nordisk's semaglutide franchise.",
        "triplets": [["Eli Lilly", "competing with", "Novo Nordisk"], ["Tirzepatide", "is competing with", "Semaglutide"]]
    },
    "f2": {
        "fact": "Eli Lilly is trying to show tirzepatide, the dual GIP/GLP-1 agonist, can strike out beyond diabetes and obesity.",
        "triplets": [["Eli Lilly", "is trying to show", "Tirzepatide"], ["Tirzepatide", "is a", "dual GIP/GLP-1 agonist"], 
                     ["Tirzepatide", "can treat beyond", "Diabetes"], ["Tirzepatide", "can treat beyond", "Obesity"]]
    },
    "f3": {
        "fact": "Tirzepatide, under the brand name Mounjaro, received its first FDA approval for Type 2 diabetes in May 2022.",
        "triplets": [["Tirzepatide", "branded as", "Mounjaro"], ["Mounjaro", "won", "FDA approval"], 
                     ["FDA approval", "for",  "Type 2 diabetes"], ["FDA approval", "was in", "May 2022"]]
    },
    "f4": {
        "fact": "Tirzepatide, under the brand name Zepbound, received an obesity approval in November 2022.",
        "triplets": [["Tirzepatide", "was branded as", "Zepbound"], ["Zepbound", "received", "Obesity approval"], 
                     ["Obesity approval", "was in", "November 2022"]]
    },
    "f5": {
        "fact": "Mounjaro generated $2.2 billion in sales in the fourth quarter of 2023, an eight-fold increase from the $279 million 
        during the same period in 2022.",
        "triplets": [["Mounjaro", "2023's fourth quarter sales", "$2.2 billion sales"], 
        ["Mounjaro", "2022's fourth quarter sales", "$279 million"]]
    },
    "f6": {
        "fact": "Mounjaro brought in around $5.2 billion in revenues year-to-date in 2023, Lilly said in an earnings release Tuesday",
        "triplets": [["Mounjaro", "2023 sales year-to-date", "$5.2 billion revenues"]]
    },
    "f7": {
        "fact": "Zepbound generated $175.8 million in sales in its first quarter on the market.",
        "triplets": [["Zepbound", "first quarter sales", "$175.8 million"]]
    },
    "f8": {
        "fact": "Eli Lilly's fourth-quarter sales were around $9.4 billion, a 28% increase over the $7.3 billion during the same 
        period in 2022.",
        "triplets": [["Eli Lilly", "2023 fourth-quarter sales", "$9.4 billion,"], 
        ["Eli Lilly", "2022 fourth-quarter sales", "$7.3 billion,"]]
    }
}
\end{verbatim}
\end{mdframed}
\caption{Prompt for extracting propositions (\emph{e.g.}, facts) and their corresponding entities from a document.}
\label{fig:prop-prompt}
\end{figure*}

\end{document}